\documentclass[11pt,letterpaper]{article}

\usepackage[hyperref]{emnlp-ijcnlp-2019}
\usepackage{times}
\usepackage{latexsym}

\usepackage{url}
\usepackage{graphicx}
\usepackage{booktabs}
\usepackage{multirow}

\usepackage{times}
\usepackage{latexsym}
\usepackage{microtype}
\usepackage{booktabs}
\usepackage{url}
\usepackage{eurosym}
\usepackage{todonotes}
\usepackage{multirow}
\usepackage{subcaption}
\usepackage{amssymb}
\usepackage{amsmath}
\usepackage{array}
\usepackage{xcolor}
\usepackage{placeins}
\usepackage{float}
\aclfinalcopy 

\newcommand\dataset{\textsc{\textbf{ROPES}}}

\newcommand{\PreserveBackslash}[1]{\let\temp=\\#1\let\\=\temp}
\newcolumntype{C}[1]{>{\PreserveBackslash\centering}p{#1}}
\newcolumntype{R}[1]{>{\PreserveBackslash\raggedleft}p{#1}}
\newcolumntype{L}[1]{>{\PreserveBackslash\raggedright}p{#1}}

\title{Reasoning Over Paragraph Effects in Situations}

\author{Kevin Lin, Oyvind Tafjord, Peter Clark, {\normalfont and} Matt Gardner \\
  Allen Institute for Artificial Intelligence \\
  {\tt \{kevinl,oyvindt,peterc,mattg\}@allenai.org}}

\date{}

\begin{document}
\maketitle
\begin{abstract}
  A key component of successfully reading a passage of text is the ability to apply knowledge gained from the passage to a new situation. In order to facilitate progress on this kind of reading, we present \dataset{}, a challenging benchmark for reading comprehension targeting \textbf{R}easoning \textbf{O}ver \textbf{P}aragraph \textbf{E}ffects in \textbf{S}ituations. We target expository language describing causes and effects (e.g., ``animal pollinators increase efficiency of fertilization in flowers''), as they have clear implications for new situations. A system is presented a background passage containing at least one of these relations, a novel situation that uses this background, and questions that require reasoning about effects of the relationships in the background passage in the context of the situation. We collect background passages from science textbooks and Wikipedia that contain such phenomena, and ask crowd workers to author situations, questions, and answers, resulting in a 14,322 question dataset. We analyze the challenges of this task and evaluate the performance of state-of-the-art reading comprehension models. The best model performs only slightly better than randomly guessing an answer of the correct type, at 61.6\% F1, well below the human performance of 89.0\%.
\end{abstract}

\section{Introduction}

\begin{figure}[t]
\centering
\fbox{
\begin{minipage}{0.44\textwidth}
\small
\textbf{Background:} Scientists think that the earliest flowers attracted {\bf \color{teal} insects and other animals, which spread pollen} from flower to flower. {\bf \color{brown} This greatly increased the efficiency of fertilization} {\bf \color{teal} over wind-spread pollen}, which might or might not actually land on another flower. {\bf \color{violet} To take better advantage of this “animal labor,” plants evolved traits such as brightly colored petals to attract pollinators}. In exchange for pollination, flowers gave the pollinators nectar.\\

\textbf{Situation:} Last week, John visited the national park near his city. He saw many flowers. His guide explained him that there are two categories of flowers, category A and category B. {\bf \color{teal} Category A flowers spread pollen via wind, and category B flowers spread pollen via animals}.\\

\textbf{Question:} Would category B flower have {\bf \color{brown} more or less efficient fertilization} than category A flower?\\
\textbf{Answer:} more\\

\textbf{Question:} Would category A flower have {\bf \color{brown} more or less efficient fertilization} than category B flower?\\
\textbf{Answer:} less\\

\textbf{Question:} Which category of flowers would be more likely to have {\bf \color{violet} brightly colored petals}? 

\textbf{Answer:} Category B\\

\textbf{Question:} Which category of flowers would be less likely to have {\bf \color{violet} brightly colored petals}? 

\textbf{Answer:} Category A\\

\end{minipage}
}  
\caption{\label{figure:dataset-example}  Example questions in \dataset{}.}
\end{figure}

Comprehending a passage of text requires being able to understand the implications of the passage on other text that is read. For example, after reading a background passage about how animal pollinators increase the efficiency of fertilization in flowers, a human can easily deduce that given two types of flowers, one that attracts animal pollinators and one that does not, the former is likely to have a higher efficiency in fertilization (Figure \ref{figure:dataset-example}). This kind of reasoning however, is still challenging for state-of-the-art reading comprehension models. Recent work in reading comprehension has seen impressive results, with models reaching human performance on well-established datasets \cite{devlin2018bert, wang2017gated,chen2016thorough}, but so far has mostly focused on extracting local predicate-argument structure, without the need to apply what was read to outside context. 

We introduce \dataset{}\footnote{https://allennlp.org/ropes}, a reading comprehension challenge that focuses on understanding causes and effects in an expository paragraph, requiring systems to apply this understanding to novel situations. If a new situation describes an occurrence of the cause, then the system should be able to reason over the effects if it has properly understood the background passage.

We constructed \dataset{} by first collecting background passages from science textbooks and Wikipedia articles that describe causal relationships. We showed these paragraphs to crowd workers and asked them to write situations that involve the relationships found in the background passage, and questions that connect the situation and the background using the causal relationships. The answers are spans from either the situation or the question. The dataset consists of 14,322 questions from various domains, mostly in science and economics.

In analyzing the data, we find (1) that there are a variety of cause / effect relationship types described; (2) that there is a wide range of difficulties in matching the descriptions of these phenomena between the background, situation, and question; and (3) that there are several distinct kinds of reasoning over causes and effects that appear.

To establish baseline performance on this dataset, we use a reading comprehension model based on RoBERTa~\citep{liu2019roberta}, reaching an accuracy of 61.6\% $F_1$. Most questions are designed to have two sensible answer choices (eg. ``more'' vs. ``less''), so this performance is little better than randomly picking one of the choices. Expert humans achieved an average of 89.0\% $F_1$ on a random sample.
\section{Related Work}
\paragraph{Reading comprehension} 
There are many reading comprehension datasets~\cite{Richardson2013MCTestAC,rajpurkar2016squad,Kwiatkowski2019NQ,Dua2019DROPAR}, the majority of which principally require understanding local predicate-argument structure in a passage of text. The success of recent models suggests that machines are becoming capable of this level of understanding. \dataset{} challenges reading comprehension models to handle more difficult phenomena: understanding the \emph{implications} of a passage of text. \dataset{} is also particularly related to datasets focusing on ``multi-hop reasoning'' \cite{Yang2018HotpotQAAD,Khashabi2018LookingBT}, as by construction answering questions in \dataset{} requires connecting information from multiple parts of a given passage.

The most closely related datasets to \dataset{} are ShARC \cite{saeidi2018interpretation}, OpenBookQA \cite{mihaylov2018can}, and QuaRel \cite{tafjord2018quarel}. ShARC shares the same goal of understanding causes and effects (in terms of specified rules), but frames it as a dialogue where the system has to also generate questions to gain complete information. OpenBookQA, similar to \dataset{}, requires reading scientific facts, but it is focused on a \emph{retrieval} problem where a system must find the right fact for a question (and some additional common sense fact), whereas \dataset{} targets \emph{reading} a given, complex passage of text, with no retrieval involved.  QuaRel is also focused on reasoning about situational effects in a question-answering setting, but the ``causes'' are all pre-specified, not read from a background passage, so the setting is limited.

\paragraph{Recognizing textual entailment} The application of causes and effects to new situations has a strong connection to notions of entailment---\dataset{} tries to get systems to understand what is entailed by an expository paragraph. The setup is fundamentally different, however: instead of giving systems pairs of sentences to classify as entailed or not, as in the traditional formulation \citep[\emph{inter alia}]{Dagan2006ThePR,Bowman2015ALA}, we give systems questions whose answers require understanding the entailment.
\section{Data Collection}
\textbf{Background passages}: We automatically scraped passages from science textbooks\footnote{We used life science and physical science concepts from www.ck12.org, and biology, chemistry, physics, earth science, anatomy and physiology textbooks from openstax.org} and Wikipedia that contained causal connectives eg. "causes," "leads to," and keywords that signal qualitative relations, e.g. "increases," "decreases."\footnote{We scraped Wikipedia online in March and April 2019}. We then manually filtered out the passages that do not have at least one relation. The passages can be categorized into physical science (49\%), life science (45\%), economics (5\%) and other (1\%). In total, we collected over 1,000 background passages.

\textbf{Crowdsourcing questions} We used Amazon Mechanical Turk (AMT) to generate the situations, questions, and answers. The AMT workers were given background passages and asked to write situations that involved the relation(s) in the background passage. The AMT workers then authored questions about the situation that required both the background and the situation to answer. In each human intelligence task (HIT), AMT workers are given 5 background passages to select from and are asked to create a total of 10 questions. To mitigate the potential for easy lexical shortcuts in the dataset, the workers were encouraged via instructions to write questions in \emph{minimal pairs}, where a very small change in the question results in a different answer. Two examples of these pairs are given in Figure \ref{figure:dataset-example}: switching ``more'' to ``less'' results in the opposite flower being the correct answer to the question.

\begin{table}[H]
\centering
\footnotesize
\begin{tabular}{lrrr}
\toprule 
 \textbf{Statistic} & \textbf{Train}  & \textbf{Dev} & \textbf{Test}\\  \midrule

\# of annotators       &     7       &     2   &   2 \\   
\# of situations       &   1411         &  203   &  300  \\   
\# of questions       &   10924       &    1688   &  1710 \\   

\midrule

avg. background length      &    121.6        &   90.7    &  123.1  \\ 
avg. situation  length     &   49.1        &     63.4 &   55.6  \\  
avg. question length     &   10.9         &      12.4  &  10.6  \\  
avg. answer length       &   1.3        &      1.4   &  1.4 \\   

\midrule

background vocabulary size       &   8616         &      2008   &  3988 \\   
situation vocabulary size       &   6949         &      1077   &   2736 \\   
question vocabulary size       &   1457        &      1411  &   1885 \\   

\bottomrule
\end{tabular}
\caption{Key statistics of \dataset{}. In total there were 588 background passages selected by the workers.}
\label{table:key_statistics}
\end{table}

\begin{table}[H]
\centering
\footnotesize
\begin{tabular}{p{1.0cm}p{5.5cm}}
\toprule 
    \textbf{Type} & \textbf{Background}  \\  \midrule
   \textbf{C (70\%)}      &     Scientists think that the earliest flowers {\bf \color{teal}{attracted insects and other animals}}, which spread pollen from flower to flower. This greatly {\bf \color{brown}{increased the efficiency of fertilization}} over wind-spread pollen ...  \\  
   \textbf{Q (4\%)}   &  ... As {\bf \color{violet}{decibel levels get higher}}, {\bf \color{violet}{sound waves have greater intensity}} and {\bf \color{violet}{sounds are louder}}. ...   \\  
   \textbf{C}\&\textbf{Q (26\%)}     &     ... Predators can be keystone species . These are species that can have a large effect on the balance of organisms in an ecosystem. For example, if all of {\bf \color{teal}{the wolves are removed from a population}}, then {\bf \color{brown}{the population of deer or rabbits may increase}}...     \\   \bottomrule
\end{tabular}
\caption{ Types of relations in the background passages. \textbf{C} refers to causal relations and \textbf{Q} refers to qualitative relations.}
\label{table:background}
\end{table}

\section{Dataset Analysis}

We qualitatively and quantitatively analyze the phenomena that occur in \dataset{}. Table \ref{table:key_statistics} shows the key statistics of the dataset. We randomly sample 100 questions and analyze the type of relation in the background, grounding in the situation, and reasoning required to answer the question.

\paragraph{Background passages} We manually annotate whether the relation in the background passage being asked about is causal (a clear cause and effect in the background), qualitative (e.g., as X increases, Y decreases), or both. Table \ref{table:background} shows the breakdown of the kinds of relations in the dataset.

\begin{table}
\centering
\footnotesize
\begin{tabular}{p{1.2cm}p{2.9cm}p{2.4cm}}
\toprule 
    \textbf{Type} & \textbf{Background}  & \textbf{Situation} \\  \midrule
   \textbf{Explicit (67\%)}   &     As {\bf \color{teal}{decibel levels get higher}}, sound waves have greater intensity and sounds are louder.  & ...First, he went to stage one, where the music was playing in {\bf \color{teal}{high decibel}}.     \\  
   \textbf{Common sense (13\%)} &  ... if we want to convert a substance from a gas to a liquid or from a {\bf \color{teal}{liquid to a solid}}, we remove energy from the system   &     ... She remembered they would be needing ice so she {\bf \color{teal}{grabbed and empty ice tray and filled it}}... \\  
   \textbf{Lexical gap (20\%)}    &    ... {\bf \color{teal}{Continued exercise}} is necessary to maintain bigger, stronger muscles...     & ... Mathew goes to the gym ... does {\bf \color{teal}{very intensive workouts}}.\\   \bottomrule
\end{tabular}
\caption{ Types of grounding found in \dataset{}. }
\label{table:grounding}
\end{table}

\begin{table*}[h]
\centering
\footnotesize
\begin{tabular}{L{1.8cm}p{3.8cm}p{3.8cm}p{2.8cm}L{1.00cm}}
\toprule
{\bf Reasoning} & {\bf Background} & {\bf Situation} & {\bf Question} & {\bf Answer} \\
 \midrule
  \textbf{ Effect comparison (71\%)} & ... gas atoms change to ions that can carry an electric current. The current causes the Geiger counter to click. {\bf \color{brown}{The faster the clicks occur}}, the {\bf \color{teal}{higher the level of radiation.}}  &  ...   Location A had {\bf \color{teal}{very high radiation}}; location B had low radiation & Would location A have {\bf \color{brown} {faster}} or slower clicks than location B? &  faster \\ 
 \midrule
 \textbf{ Effect prediction (5\%)} & ... {\bf \color{teal}{Continued exercise}} is necessary to maintain {\bf \color{brown}{bigger, stronger muscles.}}  ... &  ... Mathew goes to the gym 5 times a week and {\bf \color{teal}{does very intensive workouts.}} Damen on the other hand does not go to the gym at all and lives a mostly sedentary lifestyle. & Given Mathew suffers an injury while working out and {\bf \color{teal}{cannot go to the gym for 3 months}}, will Mathews strength increase or {\bf \color{brown}{decrease}}? & decrease \\
 \midrule
 \textbf{Cause comparison (15\%)}& ...  This {\bf \color{teal}{carbon dioxide is then absorbed by the oceans}}, which {\bf \color{brown}{lowers the pH of the water}}... & The biologists found out that the Indian Ocean had a {\bf \color{brown}{lower water pH}} than it did a decade ago, and it became acidic. The water in the Arctic ocean still had a {\bf \color{brown}{neutral to basic pH}}. & Which ocean has a {\bf \color{teal}{lower content of carbon dioxide}} in its waters? & Arctic \\
 \midrule
 \textbf{Cause prediction (1\%)} & ... Conversely, if we want to convert a substance from a gas to a liquid or from a {\bf \color{brown}{liquid to a solid}}, we {\bf \color{teal}{remove energy from the system}} and decrease the temperature. ... & ... she grabbed and empty ice tray and filled it. As she walked over to the freezer ... When she checked the tray later that day the {\bf \color{brown}{ice was ready}}. & Did the freezer add or {\bf \color{teal}{remove}} energy from the water? & remove\\ 
 \midrule
 \textbf{Other (8\%)} & ... {\bf \color{brown}{Charging an object by touching it}} with another charged object is called charging by {\bf \color{teal}{conduction}}. ...  {\bf \color{teal}{induction}} allows a change in charge {\bf \color{brown}{without actually touching the charged and uncharged objects}} to each other. & ... In case A he used {\bf \color{teal}{conduction}}, and in case B he used {\bf \color{teal}{induction}}. In both cases he used same two objects. Finally, John tried to {\bf \color{brown}{charge his phone remotely}}. He called this test as {\bf \color{brown}{case C}}. & Which experiment would be less appropriate for {\bf \color{brown}{case C}}, {\bf \color{teal}{case A}} or {\bf \color{teal}{case B}}? & case A \\
 \bottomrule
\end{tabular}
\caption{Example questions and answers from \dataset{}, showing the relevant parts of the associated passage and the reasoning required to answer the question. In the last example, the situation grounds the desired outcome and asks which of two cases would achieve the desired outcome.}
\label{tab:main_examples}
\end{table*}

\paragraph{Grounding} To successfully apply the relation in the background to a situation, the system needs to be able to ground the relation to parts of the situation. To do this, the model has to either find an \textit{explicit} mention of the cause/effect from the background and associate it with some property, use a \textit{common sense fact}, or overcome a large \textit{lexical gap} to connect them. Table \ref{table:grounding} shows examples and breakdown of these three phenomena.

\paragraph{Question reasoning}
Table \ref{tab:main_examples} shows the breakdown and examples of the main types of questions by the types of reasoning required to answer them. In an \emph{effect comparison}, two entities are each associated with an occurrence or absence of the cause described in the background and the question asks to compare the effects on the two entities. Similarly, in a \emph{cause comparison}, two entities are each associated with an occurrence or absence of the effect described in the background and the question compares the causes of the occurrence or absence. In an \emph{effect prediction}, the question asks to directly predict the effect on an occurrence of the cause on an entity in the situation. Finally, in \emph{cause prediction}, the question asks to predict the cause of an occurrence of the effect on an entity in the situation. The majority of the examples are effect or cause comparison questions; these are challenging, as they require the model to ground two occurrences of causes or effects.

\paragraph{Dataset split}
In initial experiments, we found splitting the dataset based on the situations resulted in high scores due to annotator bias from prolific workers generating many examples \cite{geva2019we}. We follow their proposal and separate training set annotators from test set annotators, and find that models have difficulty generalizing to new workers.

\section{Baseline performance}
\begin{table}[h]
\centering
\footnotesize
\begin{tabular*}{0.48\textwidth}{lp{0.8cm}p{0.8cm}p{0.8cm}p{0.8cm}}
\toprule
                  & \multicolumn{2}{c}{\bf Development} & \multicolumn{2}{c}{\bf Test} \\

    & \textbf{EM}        & \textbf{F1}       & \textbf{EM}    & \textbf{F1}   \\ \midrule
RoBERTa \textsubscript{BASE}             &    38.0      &   53.5           & 35.8 &     45.5      \\ 
   \indent - background                 &   40.7       &    59.3          & 33.7 &  46.1     \\ \midrule 
RoBERTa \textsubscript{LARGE}                 &    59.7      &      70.2        &  55.4 &     61.1      \\ 

   \indent - background                 &     48.7     &        55.2      &  53.6 &     60.4      \\ 
   \indent + RACE                 &     60.1     &        73.5      &  55.5 &     61.6      \\ \midrule

   Human                 &      -    &       -      & 82.7  &     89.0      \\ \bottomrule

\end{tabular*}
\caption{Performance of baselines and human performance on the dev and test set.}
\label{table:baselines}
\end{table}
We use the RoBERTa question answering model proposed by \citet{liu2019roberta} as our baseline and concatenate the background and situation to form the passage, following their setup for SQuAD. To estimate the presence of annotation artifacts in our dataset (and as a potentially interesting future task where background reading is done up front), we also run the baseline without the background passage. Table \ref{table:baselines} presents the results for the baselines, which are significantly lower than human performance. We also experiment with first fine-tuning on RACE \cite{lai2017race} before fine-tuning on \dataset{}.

Human performance is estimated by expert human annotation on 400 random questions with the same metrics as the baselines. None of the questions share the sample background or situation to ensure that the humans do not have an unfair advantage over the model by using knowledge of how the dataset is constructed, e.g. the fact that pairs of questions like in Table \ref{figure:dataset-example} will have opposite answers.

\section{Conclusion}
We present \dataset{}, a new reading comprehension benchmark containing 14,322 questions, which aims to test the ability of systems to apply knowledge from reading text in a new setting. We hope that \dataset{} will aide efforts in tying language and reasoning together for more comprehensive understanding of text.

\section{Acknowledgements}
We thank the anonymous reviewers for their feedback. We also thank Dheeru Dua and Nelson Liu for their assistance with the crowdsourcing setup, and Kaj Bostrom for the human evaluation. We are grateful for discussion with other AllenNLP and Aristo team members and the infrastructure provided by the Beaker team. Computations on beaker.org were supported in part by credits from Google Cloud.

\bibliography{bib}
\bibliographystyle{acl_natbib}

\appendix
\end{document}